\documentclass[11pt]{article}
\usepackage{acl2016}
\usepackage{times}
\usepackage{latexsym}
\usepackage{booktabs}
\usepackage{url}
\usepackage{graphicx}
\usepackage{acronym}
\usepackage[lined, ruled]{algorithm2e}
\usepackage[normalem]{ulem}
\usepackage{commath}
\usepackage{amsmath}
\usepackage{amssymb}
\usepackage{booktabs}

\usepackage{regexpatch,fancyvrb,xparse}
\makeatletter
\let\do@footnotetext\@footnotetext
\regexpatchcmd{\do@footnotetext}
  {\c{insert}\c{footins}\cB.(.*)\cE.}
  {\1\c{egroup}}
  {}{}
\def\@footnotetext{\insert\footins\bgroup\@makeother\#\do@footnotetext}
\newcommand{\ttvar}{\begingroup\@makeother\#\@ttvar}
\newcommand{\@ttvar}[1]{\ttfamily\detokenize{#1}\endgroup}
\makeatother

\useunder{\uline}{\ul}{}

\newcommand{\jst}{JBT }
\newcommand{\wv}{w2v }

\aclfinalcopy 
\def\aclpaperid{***} 

\DeclareMathOperator*{\argmax}{arg\,max}

\newenvironment{itemize2}
    {\begin{itemize}
        \vspace{-0.3em}
        \setlength{\abovedisplayskip}{0pt}
        \setlength{\belowdisplayskip}{0pt}
        \setlength{\itemsep}{5pt}
        \setlength{\parskip}{0pt}
        \setlength{\parsep}{0pt}
        \setlength{\topsep}{0pt}
        \setlength{\partopsep}{0pt}
    }
    {\vspace{-0.3em}
    \end{itemize}}

\title{Making Sense of Word Embeddings}

\author{Maria Pelevina$^1$, Nikolay Arefyev$^{2}$,
       Chris Biemann$^1$ and Alexander Panchenko$^1$
       \\ \\
       $^1$Technische Universit{\"a}t Darmstadt, LT Group, Computer Science Department, Germany\\
       $^2$Moscow State University, Faculty of Computational Mathematics and Cybernetics, Russia \\
       {\tt  panchenko@lt.informatik.tu-darmstadt.de} }
       
\date{}

\begin{document}
\maketitle
\begin{abstract}
 
We present a simple yet effective approach for learning word sense embeddings.
In contrast to existing techniques, which either directly learn sense representations from corpora or rely on sense inventories from lexical resources, our approach can induce a sense inventory from existing word embeddings via clustering of ego-networks of related words. An integrated WSD mechanism enables labeling of words in context with learned sense vectors, which gives rise to downstream applications. Experiments show that the performance of our method is comparable to state-of-the-art unsupervised WSD systems.  

\end{abstract}

\section{Introduction}

Term representations in the form of dense vectors are useful for many natural language processing applications. First of all, they enable the computation of semantically related words. Besides, they can be used to represent other linguistic units, such as phrases and short texts, reducing the inherent sparsity of traditional vector-space representations~\cite{salton1975vector}.

One limitation of most word vector models, including  sparse~\cite{baroni2010distributional} and dense~\cite{mikolov2013efficient} representations, is that they conflate all senses of a word into a single vector. Several architectures for learning multi-prototype embeddings were proposed that try to address this shortcoming~\cite{Huang2012,tianEtAl2014,neelakantanefficient,nietopina-johansson:2015:RANLP2015,bartunov2015breaking}. Li and Jurafsky~\shortcite{li2015multi} provide indications that such sense vectors improve the performance of text processing applications, such as part-of-speech tagging and semantic relation identification.

The contribution of this paper is a novel method for learning word sense vectors. In contrast to previously proposed methods, our approach
relies on existing single-prototype word embeddings, transforming them to sense vectors via ego-network clustering. An ego network consists of a single node (ego) together with the nodes they are connected to (alters) and all the edges among those alters. Our method is fitted with a word sense disambiguation (WSD) mechanism, and thus words in context can be mapped to these sense representations. An advantage of our method is that one can use existing word embeddings and/or existing word sense inventories to build sense embeddings. Experiments show that our approach performs comparably to  state-of-the-art unsupervised WSD systems. 



\begin{figure*}
\begin{center}
\includegraphics[width=0.75\textwidth]{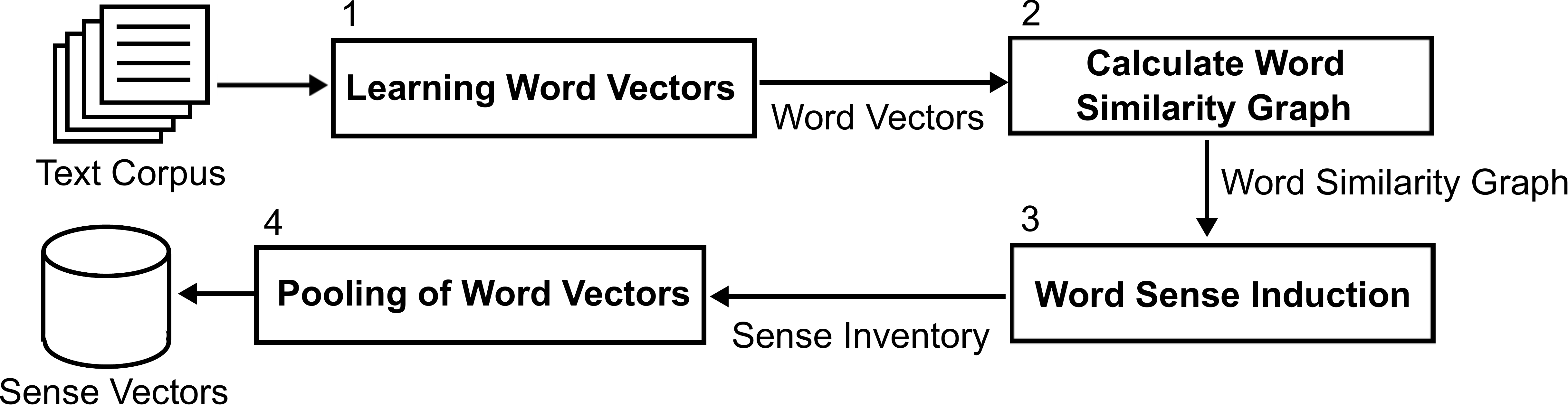}
\end{center}
\caption{Schema of the word sense embeddings learning method.}
\label{fig:pipeline}
\end{figure*}

\section{Related Work}
\label{sec:relatedwork}

Our method learns multi-prototype word embeddings and applies them to WSD. Below we briefly review both strains of research. 

\subsection{Multi-Prototype Word Vector Spaces}

In his pioneering work, Sch\"utze~\shortcite{schutze1998automatic} induced sparse sense vectors by clustering context vectors using the EM algorithm. This approach is fitted with a similarity-based WSD mechanism. Later, Reisinger and Mooney~\shortcite{reisinger2010multi} presented a multi-prototype vector space. Sparse TF-IDF vectors are clustered using a parametric method fixing the same number of senses for all words. Sense vectors are centroids of the clusters. 
	

While most dense word vector models represent a word with a single vector and thus conflate senses~\cite{mikolov2013efficient,pennington2014glove}, there are several approaches that produce word sense embeddings. Huang et al.~\shortcite{Huang2012} learn dense vector spaces with neural networks. First,  contexts are represented with word embeddings and clustered. Second, word occurrences are re-labeled  in the corpus according to the cluster they belong to. Finally, embeddings are re-trained on this sense-labeled terms. 
Tian et al.~\shortcite{tianEtAl2014} introduced a probabilistic extension of the Skip-gram model~\cite{mikolov2013efficient} that learns multiple sense-aware prototypes weighted by their prior probability. These models use parametric clustering algorithms that produce a fixed number of senses per word. 

Neelakantan et al.~\shortcite{neelakantanefficient} proposed a multi-sense extension of the Skip-gram model that was the first one to learn the number of senses by itself. During training, a new sense vector is allocated if the current context's similarity to existing senses is below some threshold. Li and Jurafsky~\shortcite{li2015multi} use a similar idea by integrating the Chinese Restaurant Process into the Skip-gram model. All mentioned above sense embeddings were evaluated on the contextual word similarity task, each  one improving upon previous models. 

Nieto and  Johansson~\shortcite{nietopina-johansson:2015:RANLP2015} presented another multi-prototype modification of the Skip-gram model. Their approach outperforms that of Neelakantan et al.~\shortcite{neelakantanefficient}, but requires as an input the number of senses for each word.

Li and Jurafsky~\shortcite{li2015multi} show that sense embeddings can significantly improve the performance of part-of-speech tagging, semantic relation identification and semantic relatedness tasks, but yield no improvement for named entity recognition and sentiment analysis.

Bartunov et al.~\shortcite{bartunov2015breaking} introduced AdaGram, a non-parametric method for learning sense embeddings based on a Bayesian extension of the Skip-gram model. The granularity of learned sense embeddings is controlled by the parameter $\alpha$. Comparisons of their approach to~\cite{neelakantanefficient} on three SemEval word sense induction and disambiguation datasets show the advantage of their method. For this reason, we use AdaGram as a representative of the state-of-the-art methods in our experiments.


Several approaches rely on a knowledge base (KB) to provide sense information. Bordes et al.~\shortcite{bordes2011} propose a general method to represent entities of any KB as a dense vector. Such representation helps to integrate KBs into NLP systems.
Another approach that uses sense inventories of knowledge bases was presented by Camacho-Collados et al.~\shortcite{camacho2015unified}. Rothe and Sch\"utze~\shortcite{rothe2015autoextend} combined word embeddings on the basis of WordNet synsets to obtain sense embeddings. The approach is evaluated on lexical sample tasks by adding synset embeddings as features to an existing WSD system. They used a weighted pooling similar to the one we use, but their method is not able to find new senses in a corpus. Finally, Nieto Pi\~{n}a and Johansson~\shortcite{nietopina2016} used random walks on the Swedish Wordnet to generate training data for the Skip-gram model.


\subsection{Word Sense Disambiguation (WSD)}
Many different designs of WSD systems were proposed, see \cite{agirre2007word,Navigli2009}. 
Supervised approaches use an explicitly sense-labeled training corpus to construct a model, usually building one model per target word~\cite{Lee2002,Klein2002}. 
These approaches demonstrate top performance in competitions, but require considerable amounts of sense-labeled examples. 

Knowledge-based approaches do not learn a model per target, but rather derive sense representation from information available in a lexical resource, such as WordNet. Examples of such system include~\cite{Lesk1986,banerjee2002adapted,pedersen2005maximizing,moro2014entity} 

Unsupervised WSD approaches rely neither on hand-annotated sense-labeled corpora, nor on handcrafted lexical resources. Instead, they automatically induce a sense inventory from raw corpora. Such unsupervised sense induction methods fall into two categories: context clustering, such as~\cite{pedersen1997distinguishing,schutze1998automatic,reisinger2010multi,neelakantanefficient,bartunov2015breaking} and word (ego-network) clustering, such as~\cite{lin1998information,pantel2002discovering,Widdows2002,Biemann2006,Hope2013}. Unsupervised methods use disambiguation clues from the induced sense inventory for word disambiguation. Usually, the WSD procedure is determined by the design of sense inventory. It might be the highest overlap between the instance's context words and the words of the sense cluster, as in \cite{Hope2013} or the smallest distance between context words and sense hubs in graph sense representation, as in \cite{Veronis2004}.

\begin{figure}
\begin{center}
\includegraphics[width=0.5\textwidth]{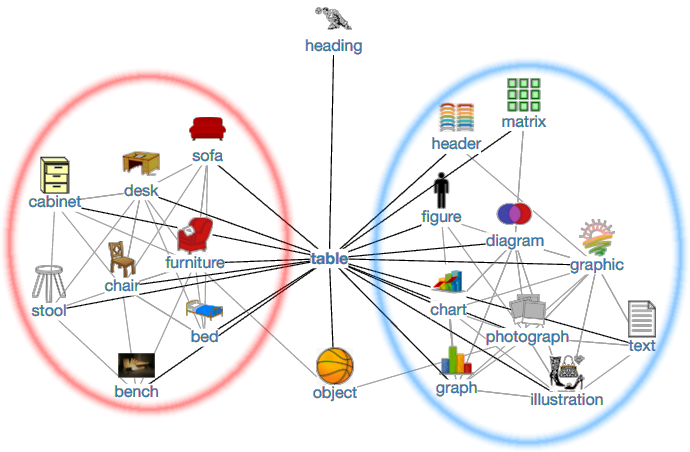}
\end{center}
\caption{Visualization of the ego-network of ``table"  with  furniture and data sense clusters. Note that the target ``table" is excluded from clustering.}
\label{graph-clustering-example}
\end{figure}

\section{Learning Word Sense Embeddings}

Our method consists of the four main stages depicted in Figure~\ref{fig:pipeline}: (1) learning word embeddings; (2) building a graph of nearest neighbours based on vector similarities; (3) induction of word senses using ego-network clustering; and (4) aggregation of word vectors with respect to the induced senses. 

Our method can use existing word embeddings, sense inventories and word similarity graphs. To demonstrate such use-cases and to study the performance of the method in different settings, as variants of the complete pipeline presented in Figure~\ref{fig:pipeline}, we experiment with two additional setups. First, we use an alternative approach to compute the word similarity graph, which relies on dependency features and is expected to provide more accurate similarities (therefore, the stage (2) is changed). Second, we use a sense inventory constructed using crowdsourcing (thus, stages (2) and (3) are skipped). Below we describe each of the stages of our method in detail.

\subsection{Learning Word Vectors}

To learn word vectors, we use the \emph{word2vec}  toolkit~\cite{mikolov2013efficient}, namely we train CBOW word embeddings with 100 or 300 dimensions, context window size of 3 and minimum word frequency of 5. We selected these parameters according to prior evaluations, e.g.~\cite{baroni2014don}, and tested them on the development dataset (see Section~\ref{twsi}). Initial experiments showed that this configuration is superior to others, e.g. the Skip-gram model, with respect to WSD performance. 

For training, we modified the standard implementation of  \textit{word2vec}\footnote{\url{https://code.google.com/p/word2vec}} so that it also saves context vectors needed for one of our WSD approaches. For experiments, we use two commonly used corpora for training distributional models: Wikipedia\footnote{We used an English Wikipedia dump of October 2015: \url{http://panchenko.me/data/joint/corpora/en59g/wikipedia.txt.gz}} and ukWaC~\cite{ferraresi2008introducing}.

\subsection{Calculating Word Similarity Graph}
\label{sec:dt-computation}

At this step, we build a graph of word similarities, such as {\small \textsf{(table, desk, 0.78)}}. For each word we retrieve its 200 nearest neighbours. This number is motivated by prior studies~\cite{Biemann2013,panchenko2013similarity}: as observed, only few  words have more strongly semantically related words. This graph is computed either based on word embeddings learned during the previous step or using semantic similarities provided by the \textit{JoBimText} framework~\cite{Biemann2013}.

\paragraph{Similarities using word2vec (w2v). } In this case, nearest neighbours of a term are terms with the highest cosine similarity of their respective vectors. For scalability reasons, we perform similarity computations via block matrix multiplications, using blocks of 1000 vectors. 

\paragraph{Similarities using JoBimText (JBT). } In this unsupervised approach, every word is represented as a bag of sparse dependency-based features extracted using the Malt parser
 and collapsed using an approach similar to~\cite{ruppert2015}. Features are normalized using the LMI score~\cite{church1990word} and further pruned down according to the recommended defaults: we keep 1000 features per word and 1000 words per feature. Similarity of two words is equal to the number of common features.

Multiple alternatives exist for computation of semantic relatedness~\cite{zhang2013recent}. JBT has two advantages in our case: (1)  accurate estimation of word similarities based on dependency features; (2) efficient computation of nearest neighbours for all words in a corpus. Besides, we observed that nearest neighbours of word embeddings often tend to belong to the dominant sense, even if minor senses have significant support in the training corpus. We wanted to test if the same problem remains for a principally different method for similarity computation.

\IncMargin{-1em}
\begin{algorithm}
\SetKwInOut{Input}{input}\SetKwInOut{Output}{output}

\Input{
$T$ -- word similarity graph, $N$ -- ego-network size, $n$ -- ego-network connectivity, $k$ -- minimum cluster size }
\Output{for each term $t \in  T$, a clustering $S_t$ of its $N$ most similar terms}
\ForEach{$t \in  T$}{	
	$V \leftarrow$ $N$ most similar terms of $t$ from $T$ \\
	$G \leftarrow$ graph with $V$ as nodes and no edges $E$
	\BlankLine
	\ForEach{$v \in V$}{
		$V' \leftarrow$ $n$ most similar terms of $v$ from $T$ \\
		\ForEach{$v' \in V'$}{
			\textbf{if} $v' \in V$ \textbf{then} add edge $(v, v')$ to $E$
		}
		
	}
	$S_t \leftarrow$ \texttt{ChineseWhispers}($G$)\\
	$S_t \leftarrow  \{s \in S_t : |s| \geq k \}$\\

}
\caption{Word sense induction.}
\label{wsi-algo-pseudocode}
\end{algorithm}

\subsection{Word Sense Induction}

We induce a sense inventory using a method similarly to \cite{pantel2002discovering} and \cite{Biemann2006}. A word sense is represented by a word cluster. For instance the cluster ``chair, bed, bench, stool, sofa, desk, cabinet" can represent the sense ``table (furniture)". To induce senses, first we construct an ego-network $G$ of a word $t$ and then perform graph clustering of this network. The identified clusters are interpreted as senses (see Table~\ref{graph-clustering-example}). Words referring to the same sense tend to be tightly connected, while having fewer connections to words referring to different senses. 

The sense induction presented in Algorithm~\ref{wsi-algo-pseudocode} processes one word $t$ of the word similarity graph $T$ per iteration. First, we retrieve nodes $V$ of the ego-network $G$: these are the $N$ most similar words of $t$ according to $T$. The target word $t$ itself is not part of the ego-network. Second, we connect the nodes in $G$ to their $n$ most similar words from $T$. 
Finally, the ego-network is clustered with the Chinese Whispers algorithm~\cite{Biemann2006}. This method is parameter free, thus we make no assumptions about the number of word senses.

The sense induction algorithm has three meta-parameters: the ego-network size ($N$) of the target ego word $t$; the ego-network connectivity ($n$) is the
maximum number of connections the neighbour $v$ is allowed to have within the ego-network; the minimum size of the cluster $k$. The $n$ parameter regulates the granularity of the inventory. In our experiments, we set the $N$ to 200, $n$ to 50, 100 or 200 and $k$ to 5 or 15 to obtain different granulates, cf.~\cite{TUD-CS-2010-23874}.  

Each word in a sense cluster has a weight which is equal to the similarity score between this word and the ambiguous word $t$.

\begin{table}
\footnotesize
\centering
\begin{tabular}{l|p{6cm}}
\toprule
\bf Vector & \bf Nearest Neighbours \\ \midrule
 table & tray, bottom, diagram, bucket, brackets, stack, basket, list, parenthesis, cup, trays, pile, playfield, bracket, pot, drop-down, cue, plate \\ \midrule
 table\ttvar{#0} & leftmost\ttvar{#}0, column\ttvar{#}1,  randomly\ttvar{#}0,  tableau\ttvar{#}1, top-left\ttvar{#}0, indent\ttvar{#}1,  bracket\ttvar{#}3,  pointer\ttvar{#}0,  footer\ttvar{#}1, cursor\ttvar{#}1, diagram\ttvar{#}0, grid\ttvar{#}0 \\ \midrule
 table\ttvar{#}1 & pile\ttvar{#}1,  stool\ttvar{#}1,  tray\ttvar{#}0,  basket\ttvar{#}0,  bowl\ttvar{#}1,  bucket\ttvar{#}0,  box\ttvar{#}0,  cage\ttvar{#}0,  saucer\ttvar{#}3,      mirror\ttvar{#}1,  birdcage\ttvar{#}0,  hole\ttvar{#}0,  pan\ttvar{#}1,  lid\ttvar{#}0 \\ 
\bottomrule
\end{tabular}
\caption{Neighbours of the word ``table" and its senses produced by our method. The neighbours of the initial vector belong to both senses, while those of sense vectors are sense-specific.   }
\label{tab:nns}
\end{table}

\begin{table*}
\footnotesize
\centering
\begin{tabular}{l|ccc}
\toprule
 & \bf \parbox{4cm}{TWSI} & \bf \parbox{4cm}{JBT} & \bf \parbox{4cm}{w2v} \\ \midrule
table (furniture) & \parbox{4cm}{counter, console, bench, dinner table, dining table, desk, surface, bar, board}  & \parbox{4cm}{chair, room, desk, pulpit, couch, furniture, fireplace, bench, door, box, railing, tray}  & \parbox{4cm}{tray, bottom, bucket, basket, cup, pile, bracket, pot, cue, plate, jar, platter, ladder} \\ \midrule
table (data) & \parbox{4cm}{chart, list, index, graph, columned list, tabulation, standings, diagram, ranking} & \parbox{4cm}{procedure, system, datum, process, mechanism, tool, method, database, calculation, scheme} & \parbox{4cm}{diagram, brackets, stack, list, parenthesis, playfield, drop-down, cube,  hash,  results, tab} \\ \midrule
table (negotiations) & \parbox{4cm}{surface, counter, console, bargaining table, platform, negotiable, machine plate, level}  & |  & | \\ \midrule
table (geo) & \parbox{4cm}{level, plateau, plain, flatland, saturation level, water table, geographical level, water level} & | & | \\ \midrule 

\end{tabular}
\caption{Word sense clusters from inventories derived from the Wikipedia corpus via crowdsourcing (TWSI), JoBimText (JBT) and word embeddings (w2v). The sense labels are introduced for readability. }
\label{tab:inventories}
\end{table*}

\subsection{Pooling of Word Vectors}

At this stage, we calculate sense embeddings for each sense in the induced inventory. We assume that a word sense is a composition of words that represent the sense. We define a sense vector as a function of word vectors representing cluster items.  Let $W$ be a set of all words in the training corpus and let $S_i = \{w_1, \dots, w_n\} \subseteq W$ be a sense cluster obtained during the previous step. Consider a function $\mathit{vec_w}: W \rightarrow \mathbb{R}^m$ that maps words to their vectors and a function $\gamma_i: W \rightarrow \mathbb{R}$ that maps cluster words to their weight in the cluster $S_i$. We experimented with two ways to calculate sense vectors: unweighted average of word vectors: 
$$
\mathbf{s}_i = \frac{\sum_{k=1}^{n} \mathit{vec_w}(w_k)}{n};
$$
and weighted average of word vectors:
$$
\mathbf{s}_i = \frac{\sum_{k=1}^{n} \gamma_i(w_k) \mathit{vec_w}(w_k)  }{ \sum_{k=1}^n \gamma_i(w_k)}.
$$

Table~\ref{tab:nns} provides an example of weighted pooling results. While the original neighbours of the word ``table" contain words related to both furniture and data, the neighbours of the sense vectors are either related to furniture or data, but not to both at the same time. Besides, each neighbour of a sense vector has a sense identifier as we calculate cosine between sense vectors, not word vectors. 

\section{Word Sense Disambiguation} 

This section describes how sense vectors are used to disambiguate a word in a context.

Given a target word $w$ and its context words $C = \{c_1, \dots, c_k\}$, we first map $w$ to a set of its sense vectors  according to the inventory: $S = \{\mathbf{s}_1, \dots, \mathbf{s}_n\}$. We use two strategies to choose a correct sense taking vectors for context words either from the matrix of context embeddings or from the matrix of word vectors. The first one is based on sense probability in given context:
$$
s^{*} = \argmax_i P(C|\mathbf{s}_i) = \argmax_i \dfrac{1}{1+e^{-\bar{\mathbf{c}}_{c} \cdot \mathbf{s}_i}},
$$ 
where  $\bar{\mathbf{c}}_{c}$ is the mean of context embeddings: $k^{-1}\sum_{i=1}^k \mathit{vec}_c(c_i)$ and functions $\mathit{vec}_c: W \rightarrow \mathbb{R}^m$ map context words to context embeddings. Using
the mean of context embeddings to calculate sense probability is natural with the CBOW because this model optimizes exactly the same mean
to have high scalar product with word embeddings for words occurred in context and low scalar product for random words~\cite{mikolov2013efficient}. 

The second disambiguation strategy is based on similarity between sense and  context:
$$
s^{*} = \argmax_i \mathit{sim}(\mathbf{s}_i, C) = \argmax_i \frac{\bar{\mathbf{c}}_{w} \cdot \mathbf{s}_i}{\norm{\bar{\mathbf{c}}_{w}} \cdot \norm{\mathbf{s}_i}},
$$
where $\bar{\mathbf{c}}_{w}$ is the mean of word embeddings: $\bar{\mathbf{c}}_{w} = k^{-1}\sum_{i=1}^k \mathit{vec}_w(c_i)$. The latter method uses only word vectors ($vec_w$) and require no context vectors ($vec_c$). This is practical, as the standard implementation of \emph{word2vec} does  not save context embeddings and thus most pre-computed models provide only word vectors.  


To improve WSD performance we also apply context filtering. Typically, only several words in context are relevant for sense disambiguation, like ``chairs" and ``kitchen" are for ``table" in ``They bought a table and chairs for kitchen." For each word $c_j$ in context $C = \{c_1, \dots, c_k\}$ we calculate a score that quantifies how well it discriminates the senses:
$$
\max_i f(\mathbf{s}_i, c_j) - \min_i f(\mathbf{s}_i, c_j),
$$
where $\mathbf{s}_i$ iterates over senses of the ambiguous word and $f$ is one of our disambiguation strategies: either $P(c_j|\mathbf{s}_i)$ or $\mathit{sim}(\mathbf{s}_i, c_j)$. The $p$ most discriminative context words are used for disambiguation.

\begin{table}
\footnotesize
\centering
\begin{tabular}{l|cc|cc}
\toprule
& \multicolumn{2}{c}{\textbf{Full TWSI}} & \multicolumn{2}{|c}{\textbf{Balanced TWSI}} \\
 & \textbf{\wv} & \textbf{\jst} & \textbf{\wv} & \textbf{\jst} \\ \midrule
no filter   & 0.676 &  0.669 & 0.383 & 0.397 \\
filter, $p$ = 5 & 0.679 & 0.674 & 0.386 & 0.403 \\
filter, $p$ = 3 & 0.681 & 0.676 & 0.387 & 0.409 \\
filter, $p$ = 2 & \bf 0.683 & \bf 0.678 & \bf 0.389 & \bf 0.410 \\
filter, $p$ = 1 & \bf 0.683 & 0.676 & \bf 0.390 & \bf 0.410 \\ \bottomrule           
\end{tabular}
\caption{Influence of context filtering on disambiguation in terms of F-score. The models were trained on Wikipedia corpus; the \wv is based on weighted pooling and similarity-based disambiguation. All differences between filtered and unfiltered models are significant ($p < 0.05$).
}
\label{tab:results-twsi-filter}
\end{table}

\section{Experiments}

We evaluate our method on two complementary datasets: (1) a crowdsourced collection of sense-labeled contexts; and (2) a commonly used SemEval dataset.   

\subsection{Evaluation on TWSI}
\label{twsi}

The goal of this evaluation is to test different configurations of our approach on a large-scale dataset, i.e. it is used for development purposes.

\begin{table*}[t!]
\footnotesize
\centering
\begin{tabular}{l|c|ccc|ccc}
\toprule
\textbf{Inventory} & \multicolumn{1}{l|}{\textbf{\#Senses}} & \multicolumn{3}{c|}{\textbf{Upper-bound of Inventory}} & \multicolumn{3}{c}{\textbf{Probability-based WSD}} \\
& \multicolumn{1}{l|}{} & \bf Prec.       & \bf Recall        & \bf F-score       & \bf Prec.       & \bf Recall       & \bf F-score      \\ \midrule
TWSI            & 2.26  & 1.000 & 1.000 &  1.000 & 0.484 & 0.483 &  0.484 \\ \midrule
\wv wiki, $k$ = 15            & 1.56  &  1.000 & 0.479 & 0.648 & 0.367 & 0.366 & 0.366 \\
\jst wiki, $n$ = 200, $k$ = 15 & 1.64  &  1.000 & 0.488 & 0.656 & \bf 0.383 & \bf 0.383 & \bf 0.383 \\ 
\jst ukWaC, $n$ = 200, $k$ = 15 & 1.89  & 1.000 & 0.526 & 0.690 & 0.360 & 0.360 & 0.360 \\
JBT wiki, $n$ = 200, $k$ = 5 & 2.55  & 1.000 & 0.598 & 0.748 & 0.338 & 0.338 & 0.338 \\
JBT wiki, $n$ = 100, $k$ = 5 & 3.59 & 1.000 & 0.671 & 0.803 & 0.305 & 0.305 & 0.305 \\
JBT wiki, $n$ = 50, $k$ = 5 & 5.13 & \bf 1.000 & \bf 0.724 & \bf 0.840 & 0.275 & 0.275 & 0.275 \\ \bottomrule
       
\end{tabular}
\caption{Upper-bound and actual value of the WSD performance on the sense-balanced TWSI dataset, function of sense inventory used for unweighted pooling of word vectors.}
\label{tab:results-twsi-inv}
\end{table*}

\paragraph{Dataset.} This test collection is based on a large-scale crowdsourced resource by Biemann~\shortcite{Biemann2012} that comprises 1,012 frequent nouns with average polysemy of 2.26 senses per word. For these nouns the dataset provides 145,140 annotated sentences sampled from Wikipedia. Besides, it is accompanied by an explicit sense inventory, where each sense is represented with a list of words that can substitute target noun in a given sentence. 

The sense distribution across sentences in the dataset is skewed, resulting in 79\% of contexts assigned to the most frequent senses. Therefore, in addition to the full TWSI dataset, we also use a balanced subset that has no bias towards the Most Frequent Sense (MFS). This dataset features 6,165 contexts with five contexts per sense excluding monosemous words. 

\paragraph{Evaluation metrics.}
To compute WSD performance,
we create an explicit mapping between the
system-provided sense inventory and the TWSI
senses: senses are represented as bag of words
vectors, which are compared using cosine similarity.
Every induced sense gets assigned to at most
one TWSI sense. Once the mapping is completed,
we can calculate precision and recall of sense prediction
with respect to the original TWSI labeling.

Performance of a disambiguation model depends on quality of the sense mapping. These baselines facilitate interpretation of results: 
\begin{itemize2}
\item \textbf{Upper bound of the induced inventory} selects the correct sense for the context, but only if the mapping exist for this sense. 
\item \textbf{MFS of the TWSI inventory} assigns the most frequent sense in the TWSI dataset. 
\item \textbf{MFS of the induced inventory} assigns the identifier of the largest sense cluster.
\item \textbf{Random sense baseline} of the TWSI and induced sense inventories.
\end{itemize2}

\paragraph{Discussion of results.}

\begin{figure*}
\begin{center}
\includegraphics[width=1.0\textwidth]{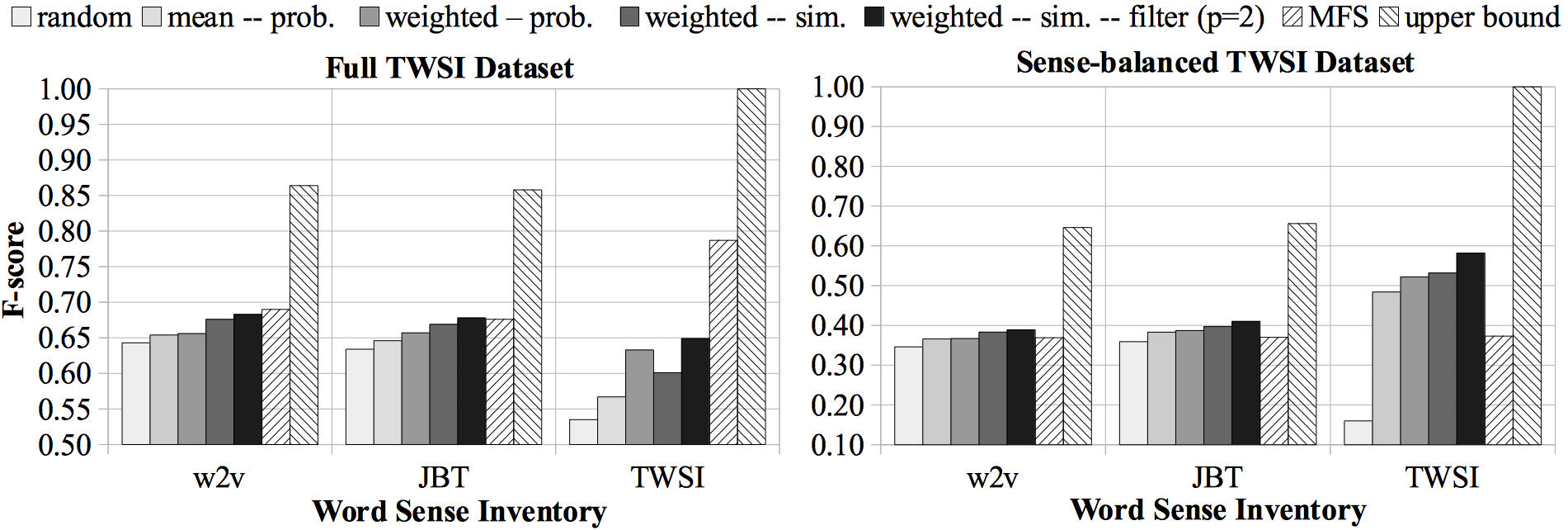}
\end{center}
\caption{WSD performance of our method trained on the Wikipedia corpus on the full (on the left) and on the sense-balanced (on the right) TWSI dataset. The \wv  models are based on the CBOW with 300 dimensions and context window size 3. The JBT models  are computed using the Malt parser. 
}

\label{fig:results-twsi}
\end{figure*}

Table~\ref{tab:inventories} presents examples
of the senses induced via clustering of nearest
neighbours generated by word embeddings (w2v)
and JBT as compared to the inventory produced
via crowdsourcing (TWSI). The TWSI contains
more senses (2.26 on average), while induced ones
have less senses (1.56 and 1.64, respectively). The
senses in the table are arranged in the way they are
mapped to TWSI during evaluation. 

Table~\ref{tab:results-twsi-filter} illustrates how the granularity of the inventory influences WSD performance. The more granular the sense inventory, the better the match between the TWSI and the induced inventory can be established (mind that we map every induced sense to at most one TWSI sense). Therefore, the upper bound of WSD performance is maximal for the most fine-grained inventories. 

However, the relation of actual WSD performance to granularity is inverse: the lower the number of senses, the higher the WSD performance (in the limit, we converge to the strong MFS baseline). We select a coarse-grained inventory for our further experiments ($n$=200, $k$ = 15). 

Table~\ref{tab:results-twsi-inv} illustrates the fact that using context filtering positively impacts disambiguation performance, reaching  optimal characteristics when two context words are used. 

Finally, Figure~\ref{fig:results-twsi} presents results of our experiments on the full and sense-balanced TWSI datasets. First of all, our models significantly outperform random sense baseline of both TWSI and induced inventories. Secondly, we observe that pooling vectors using similarity scores as weights is better than unweighted pooling. Indeed, some clusters may contain irrelevant words and thus their contribution should be discounted. Third, we observe that using similarity-based disambiguation mechanism yields better results as compared to the mechanism based on probabilities. Indeed,
cosine similarity between embeddings proved to
be useful for semantic relatedness, yielding stateof-the-art results \cite{baroni2014don}, while there is less evidence about successful use-cases of the CBOW as a language model.

Fourth, we confirm our observation that filtering context words positively impacts WSD performance. Finally, we note that models based on JBT- and w2v-induced sense inventories yield comparable results. However, the JBT inventory shows higher performance (0.410 vs 0.390) on the balanced TWSI, indicating the importance of a precise sense inventory. Finally, using the "gold" TWSI inventory significantly improves the performance on the balanced dataset outperforming models based on induced inventories.

\begin{table*}
\footnotesize

\centering
\begin{tabular}{ll|rrr|rr}
\toprule

 &  & \multicolumn{3}{c}{Supervised Evaluation} & \multicolumn{2}{c}{ Clustering Evaluation}  \\

 \bf Model &  & \bf Jacc. Ind. & \bf Tau & \bf WNDCG & \bf F.NMI & \bf F.B-Cubed \\
\midrule
Baselines & One sense for all & 0.171 & 0.627 & 0.302 & 0.000 & 0.631 \\
& One sense per instance & 0.000 & 0.953 & 0.000 & 0.072 & 0.000 \\
& Most Frequent Sense (MFS) & 0.579 & 0.583 & 0.431 &  -- & -- \\
\midrule

SemEval  & AI-KU (add1000) & 0.176 & 0.609 & 0.205 & 0.033 & 0.317 \\
& AI-KU & 0.176 & 0.619 & 0.393 & 0.066 & 0.382 \\
& AI-KU (remove5-add1000) & 0.228 & 0.654 & 0.330 & 0.040 & 0.463 \\
& Unimelb (5p) & 0.198 & 0.623 & 0.374 & 0.056 & 0.475 \\
& Unimelb (50k) & 0.198 & 0.633 & 0.384 & 0.060 & 0.494 \\
& UoS (\#WN senses) & 0.171 & 0.600 & 0.298 & 0.046 & 0.186 \\
& UoS (top-3) & 0.220 & 0.637 & 0.370 & 0.044 & 0.451 \\
& La Sapienza (1) & 0.131 & 0.544 & 0.332 & --  & -- \\
& La Sapienza (2) & 0.131 & 0.535 & 0.394 & -- & -- \\

\midrule
Sense emb. & AdaGram, $\alpha$ = 0.05, 100 dim. vectors & 0.274 & 0.644  & 0.318  & 0.058  & 0.470  \\



\toprule
Our models  & \wv   -- weighted -- sim. -- filter ($p=2$)   & 0.197 & 0.615 & 0.291 & 0.011 & 0.615 \\
            & \wv   -- weighted -- sim. -- filter ($p=2$):   nouns & 0.179 & 0.626 & 0.304 & 0.011 & 0.623 \\
            
            & \jst  -- weighted -- sim. -- filter ($p=2$)  & 0.205 & 0.624 & 0.291 & 0.017 & 0.598\\
            & \jst  -- weighted -- sim. -- filter ($p=2$): nouns & 0.198 & 0.643 & 0.310 & 0.031 & 0.595\\
            & TWSI -- weighted -- sim. -- filter ($p=2$): nouns & 0.215 & 0.651 & 0.318 & 0.030 & 0.573 \\

\bottomrule

\end{tabular}
\caption{The best configurations of our method selected on the TWSI dataset on the SemEval 2013 Task 13 dataset. The w2v-based methods rely on the CBOW model with 100 dimensions and context window size  3. The JBT similarities were computed using the Malt parser. All systems  were trained on the ukWaC corpus.}


\label{tab:results-semeval}
\end{table*}

\subsection{Evaluation on SemEval-2013 Task 13}

The goal of this evaluation is to compare the performance of our method to state-of-the-art unsupervised WSD systems.

\paragraph{Dataset.}

The SemEval-2013 task 13 ``Word Sense Induction for Graded and Non-Graded Senses"~\cite{Jurgens2013} provides 20 nouns, 20 verbs and 10 adjectives in WordNet-sense-tagged contexts. It contains 20-100 contexts per word, and 4,664 contexts in total, which were drawn from the Open American National Corpus. Participants were asked to cluster these 4,664 instances into groups, with each group corresponding to a distinct word sense. 

\paragraph{Evaluation metrics.}  Performance is measured with three measures  that require a mapping of sense inventories (Jaccard Index, Tau and WNDCG) and two cluster comparison measures (Fuzzy NMI and  Fuzzy B-Cubed). 



\paragraph{Discussion of results.}

We compare our approach
to SemEval participants and the AdaGram
sense embeddings. The \emph{AI-KU} system ~\cite{Baskaya2013} directly clusters test contexts using the $k$-means algorithm based on lexical substitution features. The \emph{Unimelb} system~\cite{Lau2013} uses a hierarchical topic model to induce and disambiguate word senses. The \emph{UoS} system \cite{Hope2013} is most similar to our approach: to induce senses it builds an ego-network of a word using dependency relations, which is subsequently clustered using a simple graph clustering algorithm. The \emph{La Sapienza} system \cite{Agirre2009}, relies on WordNet to get word senses and perform disambiguation.

Table~\ref{tab:results-semeval} shows a comparative evaluation of our method on the SemEval dataset. Like above, dependency-based (JBT) word similarities yield slightly better results than word embedding similarity (w2v) for inventory induction. In addition to these two configurations, we also built a model based on the TWSI sense inventory (only for nouns as the TWSI contains nouns only). This model significantly outperforms both JBT- and w2v-based models, thus precise sense inventories greatly influence WSD performance. 

As one may observe, performance of the best configurations of our method 
is comparable to the top-ranked SemEval participants, but is not systematically exceeding their results. AdaGram sometimes outperforms our method, sometimes it is on par, depending on the metric. We interpret these results as an indication of comparability of our method to state-of-the-art approaches. 

Finally, note that none of the unsupervised
WSD methods discussed in this paper, including
the top-ranked SemEval submissions and AdaGram,
were able to beat the most frequent sense
baselines of the respective datasets (with the exception
of the balanced version of TWSI). Similar
results are observed for other unsupervised WSD
methods~\cite{nietopina2016}.  

\section{Conclusion}

We presented a novel approach for learning of multi-prototype word embeddings. In contrast to existing approaches that learn sense embeddings directly from the corpus, our approach can operate on existing word embeddings. It can either induce or reuse a word sense inventory. Experiments on two datasets, including a SemEval challenge on word sense induction and disambiguation, show that our approach performs comparably to the state of the art. 

An implementation of our method with several pre-trained models is available online.\footnote{\url{https://github.com/tudarmstadt-lt/sensegram}}

\section*{Acknowledgments}

We acknowledge the support of the Deutsche For\-schungs\-gemeinschaft (DFG) foundation under the project "JOIN-T: Joining Ontologies and Semantics Induced from Text".

\bibliography{acl2016}
\bibliographystyle{acl2016}

\end{document}